# A deep learning experiment for semantic segmentation of overlapping characters in palimpsests

Michela Perino[1], Michele Ginolfi[2], Anna Candida Felici[3], Michela Rosellini[1]

[1] *Dipartimento di Scienze dell'antichità, Sapienza Università di Roma, Piazzale Aldo Moro, 5, 00185 Roma RM, michela.perino@uniroma1.it; michela.rosellini@uniroma1.it*
[2] *Dipartimento di Fisica e Astronomia, Università degli studi di Firenze, Via Giovanni Sansone, 1, 50019 Sesto Fiorentino FI, michele.ginolfi@unifi.it*
[3] *Dipartimento di Scienze di Base e Applicate per l'Ingegneria, Sapienza Università di Roma, Via Antonio Scarpa, 14, 00161 Roma RM, annac.felici@uniroma1.it*

*Abstract* – **Palimpsests refer to historical manuscripts where erased writings have been partially covered by the superimposition of a second writing. By employing imaging techniques, e.g., multispectral imaging, it becomes possible to identify features that are imperceptible to the naked eye, including faded and erased inks. When dealing with overlapping inks, Artificial Intelligence techniques can be utilized to disentangle complex nodes of overlapping letters. In this work, we propose deep learning-based semantic segmentation as a method for identifying and segmenting individual letters in overlapping characters. The experiment was conceived as a proof of concept, focusing on the palimpsests of the Ars Grammatica by Prisciano as a case study. Furthermore, caveats and prospects of our approach combined with multispectral imaging are also discussed.**

## I. INTRODUCTION

The term "palimpsest" originated from the practice of recycling manuscripts, which was widespread during the Middle Ages. The recycling consisted in erasing the original writing from parchment and preparing it for new writing. Palimpsests exhibit traces of two distinct and overlapping writings: the older one, erased and invisible to the naked eye, is referred to as the *scriptio inferior*, while the more recent and superimposed writing is known as the *scriptio superior*.

Currently, there is no global census of palimpsest manuscripts. However, palimpsests are not uncommon discoveries [1]. Some researchers have observed that the recycling of a manuscript was intended to make the writing material cheaper, especially in cases where the latter reached considerable costs and thus was not accessible to everyone [2, 3]. The writing surface of a palimpsest was usually made of parchment manufactured from the skins of various animals like sheep, calves, and goats [2]. Parchment was the most durable and appropriate writing support for recycling but also the most expensive.

Multispectral imaging (MSI) is one of the most effective techniques for revealing invisible traces of scraped-off writings [4, 5]. In multispectral UV imaging, the capability of enhancing the readability of text is related to the presence of ink remnants that attenuate the fluorescence emitted by the parchment support (i.e., radiation of lower energy, specifically in the visible band) [6]. The advantage of using MSI lies in the capacity of selecting the radiation coming from the object with the use of band pass filters or similar systems. However, in the case of overlapping inks, it is not always possible to select some spectral range where the "upper-text" becomes transparent, enabling to stand out the "under-text". Therefore, the letters remain partially visible preventing the full reading of the *scriptio inferior*.

Drawing from a case study involving four Latin grammar palimpsests, this work aims to explore the utilization of deep learning architectures, such as convolutional neural networks (CNNs), in combination with semantic segmentation frameworks to address the complexities arising from overlapping characters.

Semantic segmentation is a deep learning method and a fundamental task in computer vision which enables the precise identification and isolation of objects within complex scenes by assigning precise labels to each pixel. Interestingly, the task of segmenting overlapping objects is widespread across diverse domains, including astrophysics (e.g., deblending of superimposed or merging systems in large photometric images [7]; separation of kinematic components in galaxies [8]) and medical imaging (segmentation of overlapping vascular structures [9]).

Our results have demonstrated the efficacy of identifying and segmenting overlapping pairs of letters, specifically in the context of handwritings. The work serves as a proof of concept, for addressing the segmentation of overlapping letters (one belonging to *scriptio inferior*, and one belonging to *scriptio superior*) in palimpsests. This work is part of the ERC AdG 2019 PAGES (n. 882588) – "Priscian's Ars Grammatica in European Scriptoria" research project and contributes to the reading of medieval palimpsests of the Ars Grammatica by Prisciano of Cesarea.

## II. MULTISPECTRAL IMAGING OF PRISCIANO'S PALIMPSESTS

The goal of the ERC-PAGES research project is to study the Ars Grammatica written in 18 books by Prisciano in the 6th century AD [7]. It was originally conceived to teach Latin to Greek speakers, but it had a profound impact during the early Middle Ages and Renaissance. Over the centuries, due to the significant number of Greek passages in the text, the Ars stimulated Western scholars to the study of Greek language [8].

The palimpsests of the Ars consist of a group of three manuscripts in which the *scriptio inferior* ranges from the 8th to 10th centuries [9].

For multispectral imaging analysis, a monochromatic camera (QSI 6120 wsg-8) equipped with a Nikon optical system (AF-Nikkor 35Mm F/2 D) was employed, along with both Soft-Coated and Hard-Coated UV/VIS/NIR narrow-band interference filters (Thorlabs, Inc.). Figure 1 shows details of multispectral images of the three palimpsests analyzed in situ. All UV multispectral images were calibrated using dark frames with MaxIm DL Pro (Diffraction Limited). Fluorescence imaging was conducted using two UV LED lamps with emission peak at 365-370 nm (Madatec srl).

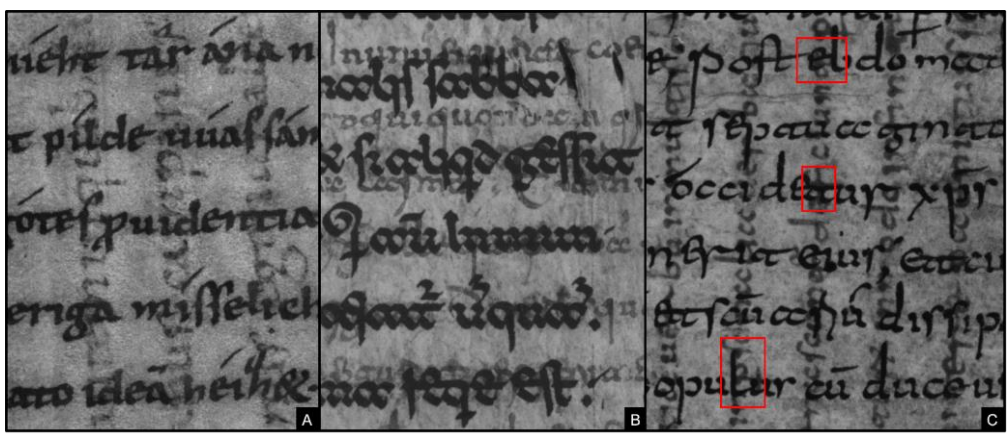

*Fig. 1. Details of multispectral images of Prisciano's palimpsests collected with UV radiation and different narrow-band filters: folio 80 from Cod. Sang. 872 kept in the Stiftsbibliothek of St. Gallen obtained with filter at 380 nm (a); folio 157r from Vallicell. C.9 kept in the Biblioteca Vallicelliana of Roma obtained with filter at 390 nm (b); folio 52r from H141 kept in the Bibliothèque Universitaire Historique de Médecine of Montpellier obtained with filter at 400 nm. Red squares enclose examples of nodes in which the overlapping writings results in indistinct characters (c).*

## III. NEURAL NETWORKS ARCHITECTURE AND TRAINING SET

This paper investigates the application of deep learning architectures, including convolutional neural networks (CNNs), in combination with semantic segmentation frameworks to effectively address challenges arising from the reading of overlapping characters.

A common architecture used for image segmentation tasks is a particular case of symmetric CNN, named encoder-decoder. The encoder extracts meaningful features from the input image, gradually reducing its spatial dimension through convolution and pooling layers. This process captures contextual information at different scales. The decoder, on the other hand, utilizes upsampling layers to reconstruct the segmented image, restoring its original size. In this study, while we tested a number of custom-made encoder-decoder models (with varying number of convolutional layers and filters), we found that generally better results are obtained through the use of a U-net architecture. The U-Net, originally introduced for biomedical image processing [13], exhibits a unique design characterized by a U-shaped topology and employs skip connections between the encoder and the decoder. Skip connections play a crucial role in preserving and integrating low-level features during the upsampling process. These connections concatenate feature maps from the encoder pathway with the feature maps at the corresponding decoder layers. By doing so, skip connections enable the U-Net architecture to recover fine-grained details and accurately localize structures in images.

As with any supervised framework, a labeled dataset is necessary to train the deep learning model. In the case of our multi-label, multi-class problem, we require a training dataset consisting of pairs input images containing overlapping letters and corresponding sets of label images containing the segmentation masks of the individual letters (prior to overlap). However, creating such a labeled dataset faces an inherent limitation: it is not always possible to accurately determine the individual letters and their strokes from the palimpsest images.

To overcome this limitation, we propose creating a synthetic dataset where input images of overlapping characters are simulated by randomly overlaying previously segmented individual letters (Figure 2).

The underlying idea is to generate a diverse and sufficiently large mock dataset that closely resembles the overwritten nodes in the palimpsest, and that can be used for training the deep learning model. The optimal approach would involve manually segmenting individual letters from both the *scriptio superior* and *inferior* in the manuscript, specifically selecting regions without overlapping writings. However, this process is time consuming, and hence not ideal to serve as a proof of concept for our proposed method.

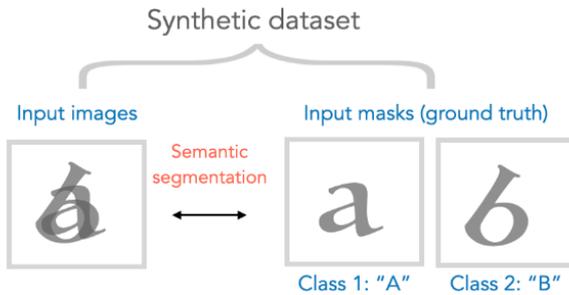

*Fig. 2. Schematic representation of the synthetic dataset used for our semantic segmentation model: input images are overlapping characters generated by randomly overlaying previously segmented individual letters (ground truth).*

To facilitate the initial exploration, we adopt a practical approach by leveraging a publicly available dataset comprising over 30,000 labeled handwritten letters (from A to Z) [14]. This dataset, predominantly sourced from the Extended MNIST database [15], contains grayscale images with dimensions of 28 x 28 pixels. For simplicity, we narrow our focus to letters A to E, encompassing five classes. While this selection does not limit the verification of our method, it streamlines the process. The dataset consists of images with a plain background, enabling us to easily generate segmentation masks for the individual letters using a simple intensity-threshold criterion. By randomly overlaying these letters, we create three synthetic datasets: one for training (150,000 entries), one for validation (3,000 entries), and one for testing (3,000 entries). Each dataset includes pairs of input images containing overlapping letters and corresponding lists of segmentation masks.

Thus, we trained a U-Net-like neural network for 15 epochs using a MacBook Pro with an M2 chip. To accelerate GPU training, we utilized TensorFlow-metal [https://www.tensorflow.org/], a framework compatible with Apple's Metal GPU. The training was conducted within the Keras framework [https://keras.io/].

## IV. RESULTS

Our model achieved a binary accuracy of approximately 97%, indicating its ability to correctly classify individual pixels. Furthermore, we obtained a precision of 93%, which measures the model's performance in avoiding false positives, and a recall of about 70%, indicating its ability to correctly identify true positives and avoid false negatives. It is important to note that while these scores are promising they should be interpreted cautiously, as they are calculated on a pixel-level basis, and in our dataset there is a higher proportion of background pixels (white pixels) compared to digits. In future developments of this study, we plan to explore more suitable metrics for evaluating model performance. However, for the present evaluation, we rely on visual inspection of the model's predictions on the test set to assess its effectiveness. In detail, for each unit of the test set, we produce a number of outputs containing (i) the input image representing either a single digit or two overlapping digits, (ii) the corresponding ground-truth labeled segmentation masks, and (iii) the predicted segmentation masks (Figure 3). To enhance visualization and explore potential patterns in noise residuals, we employ a min-max scaling technique for color-palette representation of the predicted segmentation masks. Consequently, white pixels in each panel indicate zero flux, while black pixels represent the maximum flux in the mask. Additionally, to facilitate a fair comparison between different masks and identify significant signals amidst noise residuals, we provide histograms that illustrate the distribution of maximum fluxes in each panel.

Based on visual inspection of the test set, we observed that our model successfully identified and segmented the correct combination of letters in approximately 80% of the cases. In these instances, the predicted masks for the wrong classes exhibited small flux values (below approximately 0.1 in a normalized scale). For instance, in Figure 3, the top two panels showcase accurate segmentation of complex combinations of A + C and D + E, respectively, with negligible noise residuals in the masks associated with the wrong classes.

In the remaining portion of the test set (approximately 20%), we observed two main reasons for the model's failure. In some cases (see the first panel in the bottom row of Figure 3), the model predicted significant flux in the masks associated with one or more incorrect classes. These instances mostly occurred when the random overlapping resulted in complex structures, introducing a high degree of uncertainty in the neural network's decision-making process. While these cases may not significantly improve with an augmented training set, the model's output and the distribution of predicted values in the masks could still provide useful insights for expert human classification of the letters in the node. In some other cases (see the second panel in the bottom row of Figure 3), the model displayed high confidence in predicting only two classes, but one of these predictions was incorrect. This situation likely arose due to the chance occurrence of an image resulting from the combination of two letters that resembled the morphology of a third letter. This issue can potentially be mitigated by employing a larger and more tailored training set, which would allow the model to better learn the distinguishing features of each class.

Overall, our model demonstrates promising results, validating the hypothesis that deep learning techniques hold great potential for enhancing the readability of palimpsests. The successful application of our model in accurately segmenting and identifying overlapping letters supports the notion that these advanced methods can contribute significantly to the field.

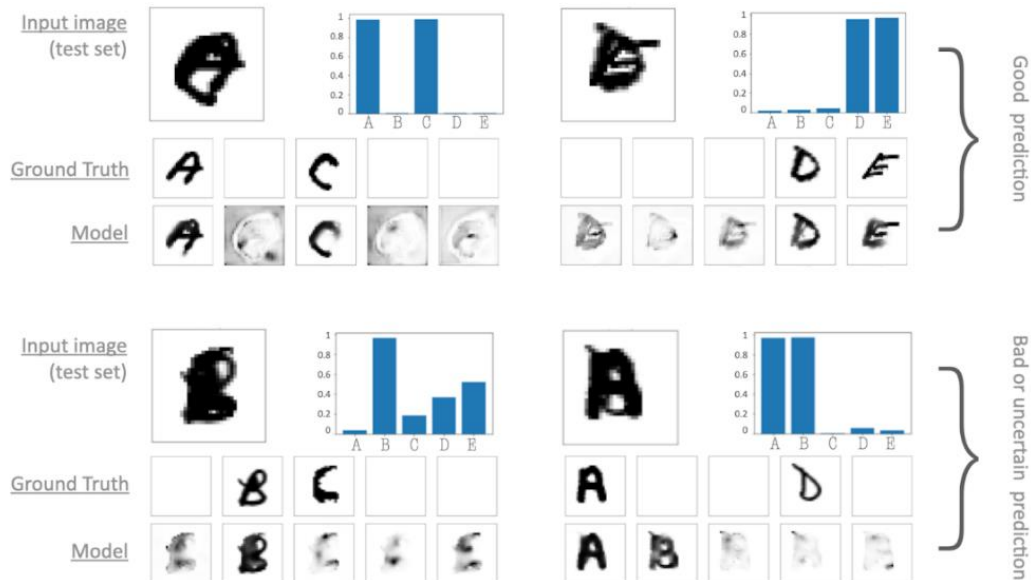

*Fig. 3. Examples of prediction of our deep learning model: in the top panels the segmentation of complex combinations of A + C and D + E; in the bottom panels the segmentation of complex combinations of B + C and A + D.*

## V. DISCUSSION AND CONCLUSIONS

In this study, we explored the potential of deep learning techniques for addressing the challenge of deciphering partially legible overlapping writings in palimpsests. We assessed the effectiveness of the semantic segmentation method in identifying and segmenting overlapping pairs of letters using a synthetic dataset. The input images of overlapping characters were generated by randomly superimposing previously segmented individual letters, primarily obtained from the Extended MNIST database. However, we acknowledged that our grayscale dataset differs from the actual historical manuscript case. To better simulate the appearance of writings in the multispectral images of palimpsests, we are currently in the process of improving our database by introducing varying contrast between the under-text and the upper-text (ongoing work). Additionally, we are testing the injection of Gaussian noise in the background. Preliminary results show that our model performances do not drop significantly due to these modifications. In the future, we plan to incorporate a more realistic simulated noise, i.e., more similar to the one visible in the multispectral images. One compelling approach to achieve this objective is by harnessing generative models like generative adversarial networks (GANs) and variational auto-encoders.

For future applications involving palimpsests, a labeled dataset is essential for training the deep learning model. Therefore, our objective is to create a synthetic dataset comprising images directly from digitized historical manuscripts. Handwriting exhibits numerous variations, not only due to different scribes but also because of changes in writing styles over the centuries. To facilitate the initial approach and evaluate the model's effectiveness, our idea is to directly segment writing from multispectral images of Prisciano's palimpsests, labelling selected portions of non-overlapping letters belonging to both *scriptio inferior* and *scriptio superior*. The input images will be then randomly superimposed, as we have described in the present work. The segmentation will be limited to palimpsests where the writing style of the *scriptio inferior* and *scriptio superior* is quite similar. Since the dataset may be too small, one way to overcome this limitation could be through artificial data augmentation, by both applying morphological transformations and leveraging generative models. The outcomes will contribute to the new digital critical edition of the Ars Grammatica and enhance our understanding of Prisciano's work.